\documentclass{article}

\usepackage[final]{neurips_2025}

\usepackage[utf8]{inputenc} 
\usepackage[T1]{fontenc}    
\usepackage{hyperref}       
\usepackage{url}            
\usepackage{booktabs}       
\usepackage{amsfonts}       
\usepackage{nicefrac}       
\usepackage{microtype}      
\usepackage{xcolor}         
\usepackage{multirow}
\usepackage{bm}
\usepackage{amsmath}
\usepackage{graphicx}
\usepackage{tikz}
\usepackage{amssymb}
\usepackage{algorithm}
\usepackage{algorithmic}
\usepackage{listings}
\usepackage{xcolor}
\usetikzlibrary{arrows.meta, positioning}

\definecolor{commentblue}{rgb}{0.25,0.5,0.75}
\title{PIRF: Physics-Informed Reward Fine-Tuning for Diffusion Models}

\author{
  Mingze Yuan \\
  Harvard University \\
  \texttt{mingzeyuan@g.harvard.edu} \\
  \And
  Pengfei Jin \\
  Massachusetts General Hospital \\
  \texttt{pjin1@mgh.harvard.edu} \\
  \And
  Na Li\thanks{Corresponding authors.} \\
  Harvard University \\
  \texttt{nali@seas.harvard.edu} \\
  \And
  Quanzheng Li\textsuperscript{*} \\
  Massachusetts General Hospital \\
  \texttt{li.quanzheng@mgh.harvard.edu}\\
}

\begin{document}

\maketitle

\begin{abstract}
Diffusion models have demonstrated strong generative capabilities across scientific domains, but often produce outputs that violate physical laws. We propose a new perspective by framing physics-informed generation as a sparse reward optimization problem, where adherence to physical constraints is treated as a reward signal. This formulation unifies prior approaches under a reward-based paradigm and reveals a shared bottleneck: reliance on diffusion posterior sampling (DPS)-style value function approximations, which introduce non-negligible errors and lead to training instability and inference inefficiency. To overcome this, we introduce Physics-Informed Reward Fine-tuning (PIRF) — a method that bypasses value approximation by computing trajectory-level rewards and backpropagating their gradients directly. However, a naive implementation suffers from low sample efficiency and compromised data fidelity. PIRF mitigates these issues through two key strategies: (1) a layer-wise truncated backpropagation method that leverages the spatiotemporally localized nature of physics-based rewards, and (2) a weight-based regularization scheme that improves efficiency over traditional distillation-based methods. Across five PDE benchmarks, PIRF consistently achieves superior physical enforcement under efficient sampling regimes, highlighting the potential of reward fine-tuning for advancing scientific generative modeling. Our code is available at \url{https://github.com/mingze-yuan/PIRF}.
\end{abstract}

\section{Introduction}
Diffusion models~\cite{ho2020denoising,song2020score,karras2022elucidating} have emerged as powerful generative tools across modalities such as images~\cite{dhariwal2021diffusion}, videos~\cite{ho2022video}, and text~\cite{nie2025large}. Recently, their use has expanded into scientific machine learning~\cite{shu2023physics,huang2024diffusionpde,li2025physicsaligned}, where the goal is to generate data governed by known physical laws—typically described by partial differential equations (PDEs). In this context, models must not only fit observed data but also satisfy physical constraints—a requirement we refer to as physical enforcement. However, standard diffusion models are trained purely with data-driven objectives and lack mechanisms to enforce such constraints, often resulting in physically invalid outputs~\cite{wang2024recent,li2024generative,qian2025physdiff}.

Existing physics-informed diffusion models tackle this challenge by incorporating physical supervision during inference and/or training. Guidance-based methods\cite{huang2024diffusionpde,jacobsen2025cocogen,shu2023physics} steer pretrained models using gradients from physics-informed losses\cite{raissi2019physics}, but typically require thousands of inference steps, making them computationally expensive. Training-based alternatives like physics-informed diffusion model (PIDM)~\cite{bastek2025physicsinformed} inject physical loss during denoising steps, improving efficiency but often at the cost of stability and performance.

We introduce a new perspective on physics-informed generation by framing it as a sparse reward optimization problem, inspired by recent work that interprets diffusion sampling as a form of sequential decision-making~\cite{uehara2024understanding,black2024training,clark2024directly}. In our formulation, rewards—derived from physical enforcement—are computed only at the final step of the diffusion trajectory. This viewpoint provides a unifying lens on prior methods: guidance-based approaches correspond to value-weighted sampling, where the gradient of an approximated value function is used to steer the inference process; meanwhile, PIDM~\cite{bastek2025physicsinformed} can be interpreted as enforcing a constant value function during training, avoiding the need to compute value gradients at inference time. However, both methods rely on diffusion posterior sampling (DPS)-style\cite{chung2022diffusion} value function approximations, which introduce significant errors in high-dimensional, non-convex PDE solution spaces\cite{li2025physicsaligned}. These errors ultimately limit the stability and efficiency.


To overcome these limitations, we propose Physics-Informed Reward Fine-Tuning (PIRF)—a method that adapts reward fine-tuning (recently applied to text-to-image generation~\cite{clark2024directly,black2024training}) to the physics-informed setting. PIRF avoids value approximation by directly computing trajectory-level rewards and backpropagating their gradients to fine-tune the model. However, a naive implementation leads to suboptimal data fidelity and inefficient sample usage. To address these issues, PIRF introduces two key innovations: First, we propose a layer-wise truncation strategy that updates only higher-resolution layers, motivated by the observation that physics-based rewards are often spatiotemporally localized. This improves training stability and data fidelity. Second, to improve sample efficiency, we replace costly distillation-based regularization with an offline weight-based regularizer, where we approximate the regularized effect using interpolated weights to greatly enhance efficiency.

We evaluate PIRF on five PDE benchmarks~\cite{shu2023physics,huang2024diffusionpde}, demonstrating consistent improvements over both guidance-based methods and PIDM in terms of physical enforcement. Notably, PIRF achieves strong performance under highly efficient inference regimes (e.g., 20 steps) and does not require gradient computations at test time. Ablation studies confirm the effectiveness of our design choices.

Our contributions are summarized as follows:
\begin{enumerate}
    \item We offer a new perspective on physics-informed generation from reward optimization and identify a common bottleneck in prior approaches: value function approximation.
    \item We introduce reward fine-tuning to the physics-informed setting and develop two strategies, layer-wise truncation and weight-based regularization.
    \item We validate PIRF on five PDE benchmarks and demonstrate state-of-the-art performance in physical enforcement under efficient sampling regimes.
\end{enumerate}

\section{Background}
\textbf{Diffusion models}~\cite{ho2020denoising,song2020score} generate data by progressively transforming a sample from a simple prior distribution—typically a standard Gaussian—into a sample drawn from the target data distribution \( q(\bm{x}) \), where \( \bm{x} \in \mathcal{X} \). This is accomplished via a fixed forward process that gradually adds Gaussian noise to a data sample \( \bm{x}_0 \sim q(\bm{x}) \) over \( T \) steps using a noise schedule \( \{ \beta_t \in (0, 1) \}_{t=1}^{T} \):
\begin{equation}
    q(\bm{x}_{1:T} | \bm{x}_0) = \prod_{t=1}^T q(\bm{x}_t | \bm{x}_{t-1}), \quad
    \text{with } q(\bm{x}_t | \bm{x}_{t-1}) = \mathcal{N}(\bm{x}_t; \sqrt{1 - \beta_t }\, \bm{x}_{t-1}, \beta_t \bm{I}).
\end{equation}
To generate new samples, a learned reverse process approximates the true posterior \( q(\bm{x}_{t-1}|\bm{x}_t) \) with a neural network parameterized by \( \theta \). Training is done by minimizing a simplified variational bound~\cite{ho2020denoising}. In this work, we use the denoising objective~\cite{karras2022elucidating}:
\begin{equation}
    \min_\theta \mathbb{E}_{\bm{x}_0 \sim q(\bm{x}), t, \boldsymbol{\epsilon} \sim \mathcal{N}(0, \bm{I})} \left\| D_\theta(\bm{x}_t, t) - \bm{x}_0 \right\|_2^2,
\end{equation}
where \( \bm{x}_t = \sqrt{\bar{\alpha}_t} \bm{x}_0 + \sqrt{1 - \bar{\alpha}_t} \boldsymbol{\epsilon} \), and \( D_\theta \) is the neural network used to predict the clean signal \( \bm{x}_0 \). 
For sampling, we start from $\bm{x}_T \sim \mathcal{N}(0, \bm{I})$ and gradually sample $\bm{x}_{t-1} \sim q(\bm{x}_{t-1}|\bm{x}_t)$. This conditional probability is not directly computable, and in practice, DDIM~\cite{song2021denoising} samples $\bm{x}_{t-1}$ via
\begin{equation}
    \bm{x}_{t-1} = \mu_\theta(\bm{x}_t, t) =\sqrt{\bar{\alpha}_{t-1}}D_\theta(\bm{x}_t, t) + \sqrt{1 - \bar{\alpha}_{t-1} - \sigma_{t}^2} \frac{\bm{x}_t - \sqrt{\bar{\alpha}_{t}}D_\theta(\bm{x}_t ,t )}{\sqrt{1 - \bar{\alpha}_t}} + \sigma_t \bm{\epsilon},
\end{equation}
where $\{\sigma_t\}_{t=1}^{T}$ are DDIM parameters, $\bm{\epsilon} \sim \mathcal{N}(0,\bm{I})$. By choosing $\sigma_t =0$, it reduces to a deterministic sampling process.

\textbf{Physical laws} are often formulated as partial differential equations (PDEs) over a domain \( \Omega = \Omega_1 \times \Omega_2 \subset \mathbb{R}^d \times \mathbb{R} \), expressed as:
\begin{equation}
\label{equ:pde}
\boldsymbol{\mathcal{G}}\big[\boldsymbol{x}(\boldsymbol{\xi})\big] = 0, \quad \boldsymbol{\xi} = (\xi_1, \xi_2, \dots, \xi_d, \tau)^\top \in \Omega,
\end{equation}
where \( \boldsymbol{\mathcal{G}} \) denotes a differential operator encompassing the boundary and/or initial conditions, and \( \boldsymbol{x}(\boldsymbol{\xi}) \in \mathbb{R}^c \) is the solution field that satisfies the set of PDEs over the domain \( \Omega \). Here, \( d \) denotes the spatial dimension and \( c \) the number of field components. The spatial domain is given by \( \Omega_1 \subset \mathbb{R}^d \), and the time domain by \( \Omega_2 \subset \mathbb{R} \). When the governing equation is time-independent, the problem reduces to a \textit{static} PDE defined solely on the spatial domain \( \Omega = \Omega_1 \), with \( \boldsymbol{\xi} = (\xi_1, \xi_2, \dots, \xi_d)^\top \). 

Note that, in general cases, the physical field can be decomposed into a solution field \( \bm{u}(\bm{\xi}) \in \mathbb{R}^{c_1}\) and a PDE coefficient field \( \bm{a}(\bm{\xi})  \in \mathbb{R}^{c_2}\)~\cite{huang2024diffusionpde}, which together characterize the physical system with $c=c_1+c_2$. That is, we define \( \bm{x} = [\bm{u}; \bm{a}] \) as the concatenation of the solution and coefficient fields. For instance, in the context of a 2D Darcy flow problem in porous media~\cite{shu2023physics,bastek2025physicsinformed}, \( \boldsymbol{x} \) describes the permeability and pressure fields, so \( c = 2 \) corresponds to these two fields, and \( d = 2 \) corresponds to the 2D spatial setting.

The \textit{PDE residual} quantifies the discrepancy between candidate fields and the governing equations. It is defined as:
\begin{equation}
    \boldsymbol{\mathcal{R}}(\bm{x}) = \boldsymbol{\mathcal{G}}[\bm{x}].
\end{equation}
For evaluation, we use the mean square error (MSE) between the PDE residual and zero, \( \operatorname{MSE}(\boldsymbol{\mathcal{R}}(\bm{x}), \bm{0}) \), as a scalar metric reflecting the degree of physical law enforcement.

\section{Methodology}

\subsection{Problem statement}
We aim to train a diffusion model whose samples comply with a set of governing PDEs, expressed as
$ \bm{\mathcal{R}}(\bm{x}) = \bm{0} $,
where $ \bm{\mathcal{R}} $ denotes the PDE residual operator. We formulate this problem from a reward optimization perspective, treating physical compliance as a reward function. The denoising process can be mapped to a Markov decision process (MDP)~\cite{fan2023dpok,black2024training} as follows (see~\autoref{fig:graphic}):
\begin{equation}
    s_j \triangleq (t, \bm{x}_t), \quad a_j \triangleq \bm{x}_{t-1}, \quad \pi(a_j | s_j) \triangleq p_t(\bm{x}_{t-1} |\bm{x}_t; \theta), \quad P(s_{j+1} | s_j, a_j) \triangleq (\delta_{t-1}, \delta_{\bm{x}_{t-1}}).
\end{equation}
Here, $ j = T - t $ is used for notational convenience, and $ \delta_y $ denotes the Dirac delta distribution centered at $ y $. The initial state distribution is defined as
$ \rho_0(s_0) = (\sigma_T, \mathcal{N}(0, \bm{I})) $.
Each trajectory consists of $ T $ steps, after which the transition dynamics $ P $ lead to a terminal state. The reward is sparse and is only defined at the final state:
$ R(s_T, a_T) = r(\bm{x}_0) $.
By defining the physics reward as the negative mean squared PDE residual, we obtain the following objective:
\begin{equation}
\label{equ:rl}
    \theta^* = \operatorname*{argmax}_{\{p_t(\bm{x}_{t-1} \mid \bm{x}_t; \theta)\}_{t=0}^{T-1}} \mathbb{E}_{\{p_t\}} \left[ r(\bm{x}_0) \right], \quad r(\bm{x}) \triangleq -\| \bm{\mathcal{R}}(\bm{x}) \|_2^2.
\end{equation}

\subsection{Casting prior works as reward-based paradigms}
\label{sec:prior}
This new perspective of reward optimization allows us to re-examine the prior physics-informed diffusion models. As summarized in~\autoref{tab:summary-reward}, Prior works incorporate the physics constraints by reshaping the inference and/or training process with reward-based paradigms. A detailed discussion of related work is provided in~\autoref{sec:related}.

Pure data-driven baselines select high-reward data from numerical simulations and train models directly on these samples. This strategy can be seen as a special case of \textit{reward-weighted regression (RWR)}, where samples are filtered using a reward threshold and used in a weighted training objective:
\begin{equation}
\label{equ:rwr}
    \min_\theta \mathbb{E}_{\bm{x}_0, t, \bm{\epsilon}} \left[ \omega(r(\bm{x}_0)) \cdot \|D_\theta(\bm{x}_t,t) - \bm{x}_0 \|_2^2 \right],
\end{equation}
where the indicator function $\omega(r) = \mathbf{1}\left( r > \eta \right)$ selects samples with reward above threshold $\eta$. We denote the resulting model as $D_{\theta_{\text{base}}}$, used as the base model for guidance-based approaches.

\textbf{Guidance-based} methods~\cite{huang2024diffusionpde,jacobsen2025cocogen} adjust the inference process to steer it toward high-reward regions. Recent work~\cite{uehara2024understanding} connects such methods to \textit{value-weighted sampling}~\cite{dhariwal2021diffusion,chung2022diffusion}, where the mean of the reverse transition is modified using the gradient of a value function:
\begin{equation}
\label{equ:valuegradient}
    \tilde{\bm{\mu}}_\theta(\bm{x}_t,t) = \bm{\mu}_{\theta}^{\text{base}}(\bm{x}_t,t) + \alpha\Sigma_t \nabla_{\bm{x}_t} v_t(\bm{x}_t),
\end{equation}
where $v_t(\bm{x}_t)$ is the value function estimating expected reward from state $\bm{x}_t$:
\begin{equation}
\label{equ:value}
    v_t(\bm{x}_t) = \mathbb{E}_{\{p_\tau\}_{\tau=t}^{1}}\left[r(\bm{x}_0) |\bm{x}_t \right].
\end{equation}
In practice, $v_t$ is approximated using a diffusion posterior sampling (DPS)-style method~\cite{chung2022diffusion}:
\begin{equation}
\label{equ:approx}
    v_t(\bm{x}_t) = \mathbb{E}[r(\bm{x}_0)| \bm{x}_t] 
    \overset{(i)}{\approx} r(\mathbb{E}[\bm{x}_0 | \bm{x}_t]) 
    \overset{(ii)}{\approx} r(D_\theta(\bm{x}_t, t)).
\end{equation}
Approximation (i) introduces a bias known as the Jensen gap~\cite{chung2022diffusion,gao2017bounds,simic2008global}, and approximation (ii) uses the denoising network as a point estimator, which can have high variance, particularly at early timesteps. These issues limit the physical precision of guidance-based methods. Moreover, the need to compute $\nabla_{\bm{x}_t} v_t(\bm{x}_t)$ slows down inference significantly.

\textbf{Training-based} methods aim to incorporate the physical constraints into the training process. For example, PG-Diffusion~\cite{shu2023physics} uses the gradient of reward as condition for classifier-free guidance (CFG)~\cite{ho2021classifier}. \textbf{PIDM}~\cite{bastek2025physicsinformed} implicitly enforces a \textit{constant value function} during training (termed as constant value forcing), avoiding computing gradients of $v_t$ in Equation~\eqref{equ:valuegradient}, thereby improving inference efficiency. Though PIDM is originally derived from virtual observables~\cite{rixner2021probabilistic}, we reinterpret it as encouraging
\begin{equation}
    v_t(\bm{x}_t) \rightarrow \max_{\bm{x} \in \mathcal{X}} r(\bm{x}) = 0 \quad \forall \bm{x}_t, t,
\end{equation}
based on the assumption that samples from numerical simulations attain the maximum reward (i.e., zero PDE residual). PIDM implements this value forcing via an augmented objective:
\begin{equation}
\label{equ:valuealign}
    \min_\theta \mathbb{E}_{\bm{x}_0, t , \bm{\epsilon}} \left[\|D_\theta(\bm{x}_t; \sigma(t)) - \bm{x}_0\|_{2}^{2} + \gamma_t \|v_t(\bm{x}_t) - 0\|_{2}^{2} \right].
\end{equation}
Since $v_t$ is not directly available, PIDM approximates it via Equation~\eqref{equ:approx}, leading to:
\begin{equation}
\label{equ:pidm}
    \min_\theta \mathbb{E}_{\bm{x}_0, t , \bm{\epsilon}} \left[\|D_\theta(\bm{x}_t; \sigma(t)) - \bm{x}_0\|_{2}^{2} + \gamma_t \|r(D_\theta(\bm{x}_t, t))\|_{2}^{2} \right],
\end{equation}
which matches the training objective in PIDM. Although a two-step DDIM estimation is proposed to reduce the error in approximation (ii) (from Equation~\eqref{equ:approx}), the bias from approximation (i) remains unaddressed. Consequently, due to approximation and optimization errors, value forcing cannot be perfectly achieved, which limits the effectiveness of the method.

\begin{table}[t]
    \centering
    \caption{Casting physics-informd diffusion models into reward-based paradigms.}
    \small
    \begin{tabular}{p{2.4cm} p{5.0cm} p{1.4cm} p{1.5cm} p{1.8cm}}
        \toprule
        Approach & 
        Reward-based \newline paradigm & 
        Inference \newline efficiency & 
        Value \newline approx. & 
        Physical \newline Enforcement \\
        \midrule
        Pure data-driven & Reward-weighted regression & High & No & Low \\
        Guidance-based   & Value-weighted sampling    & Low  & Yes & Medium \\
        PG-Diffusion~\cite{shu2023physics} & Reward gradient-conditioned CFG & Low & No & Medium \\
        PIDM~\cite{bastek2025physicsinformed}             & Constant value forcing & High & Yes & Medium \\
        Ours             & Reward backpropagation     & High & No  & High \\
        \bottomrule
    \end{tabular}
    \label{tab:summary-reward}
\end{table}

\noindent \textbf{Limitations.} 
Pure data-driven baselines lack explicit physical constraints and generally perform poorly in terms of physical enforcement. Both guidance-based approaches and PIDM rely heavily on \textit{value function approximation}, typically implemented via Equation~\eqref{equ:approx}. While this strategy is effective in some inverse problems~\cite{chung2022diffusion}, we empirically observe that it introduces substantial error in physics-informed generation—especially under efficient sampling regimes (see~\autoref{fig:value}). These limitations motivate us to sidestep value function approximation.

\begin{algorithm}[ht]
\caption{A high-level framework of reward backpropagation}
\label{alg:reward_backprop}
\begin{algorithmic}[1]
\REQUIRE Pre-trained model $\{ p_t(\bm{x}_{t-1}|\bm{x}_{t}; \theta^{\text{base}})\}_{t=0}^{N-1}$, batch size $m$, learning rate $\gamma$, physics-informed reward function $r(\cdot)$ (defined in Equation~\eqref{equ:rl}), number of iterations $K$
\STATE Initialize parameters: $\theta_1 \leftarrow \theta^{\text{base}}$
\FOR{$k = 1$ {\bfseries to} $K$}
    \STATE Sample $m$ trajectories $\{\bm{x}_t^{(i)}(\theta_k)\}_{t=0}^N$ from current model $\{ p_t(\bm{x}_{t-1}|\bm{x}_{t}; \theta_k)\}_{t=T}^{1}$
    \STATE Update parameters:
    \(
    \theta_{k+1} \leftarrow \theta_k + \gamma \nabla_\theta \left.\left[
        \frac{1}{m} \sum_{i=1}^m  
            r(\bm{x}_0^{(i)}(\theta_k)) 
    \right]\right|_{\theta = \theta_s}
    \)
\ENDFOR
\RETURN fine-tuned $\{p_t(\cdot \mid \cdot; \theta_K)\}_{t=0}^{N-1}$
\end{algorithmic}
\end{algorithm}
\subsection{PIRF: Physics-Informed Reward Fine-Tuning}

Motivated by recent work on text-to-image diffusion alignment~\cite{clark2024directly,wu2024deep}, we introduce reward fine-tuning into the physics-informed setting, termed as Physics-Informed Reward Fine-Tuning (PIRF). 

\begin{figure}[t]
    \centering
    \includegraphics[width=\linewidth]{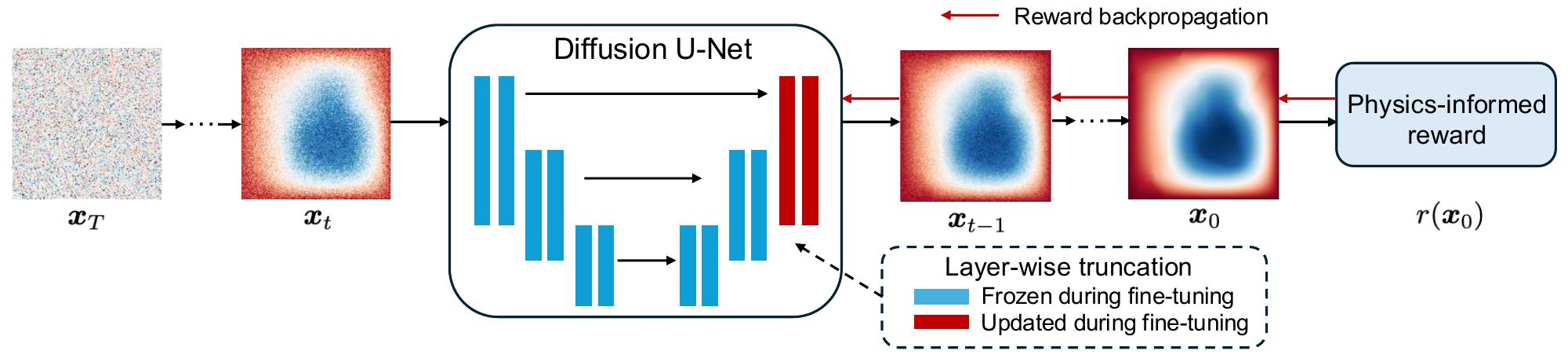}
    \caption{PIRF bypasses the challenges of value function approximation by directly backpropagating gradients from accurate final-state rewards. The proposed layer-wise truncation strategy, which updates only high-resolution layers, further enhances training stability.}
    \label{fig:method}
\end{figure}

\noindent\textbf{Overview.} As shown in~\autoref{fig:method} and Algorithm~\ref{alg:reward_backprop}, PIRF starts with a pre-trained model $\theta_{\text{base}}$ via standard diffusion training~\cite{karras2022elucidating}. During fine-tuning, it bypasses value function approximation by iterating over two phases: (1) sampling full diffusion trajectories (line 3), (2) fine-tuning the model using accurate reward signals (line 4). Since physics-based rewards are differentiable, we directly backpropagate gradients of the final-step reward $r(\bm{x}_0)$ through the entire denoising trajectory~\cite{clark2024directly}.

For notational simplicity, we formulate it with the deterministic sampling setting, where trajectories begin from $\bm{x}_T \sim \mathcal{N}(0, \bm{I})$ and evolve as:
\begin{equation}
    \bm{x}_{t-1} = \mathcal{F}_\theta^{(t)}(\bm{x}_t) \triangleq \bm{\mu}_\theta(\bm{x}_t, t), \quad t = 1, \cdots, T,
\end{equation}
leading to a final state:
\begin{equation}
    \bm{x}_0 = \mathcal{F}_\theta^{(1)} \circ \cdots \circ \mathcal{F}_\theta^{(T)}(\bm{x}_T).
\end{equation}

While conceptually straightforward, this approach is sample inefficient as it requires computing full trajectories for every update. Additionally, it introduces two major challenges: (1) memory overhead from storing all intermediate states during backpropagation; and (2) \textit{reward hacking}, where the model sacrifices data fidelity or diversity to over-optimize the reward. These challenges motivate the need for practical improvements.

\noindent \textbf{Step-wise truncation baseline.}
Prior work such as DRaFT~\cite{clark2024directly} addresses memory inefficiency via step-wise truncation, limiting backpropagation to the last $K$ steps. PalSB~\cite{li2025physicsaligned} extends this to a physics-aligned Schrödinger bridge model. We adopt this approach as a baseline (termed PIRF-base), but observe that in physics-informed generation, step-wise truncation still allows excessive model flexibility, leading to overfitting and reward hacking.

\noindent \textbf{Layer-wise truncation (LT).} 
To further constrain optimization, we introduce a finer-grained truncation strategy at the network layer level. Specifically, the denoising operator $\mathcal{F}_{\theta}^{(t)}(\bm{x}_t)$ is composed of $M$ neural network layers:
\begin{equation}
    \mathcal{F}_{\theta}^{(t)}(\bm{x}_t) = \mathcal{F}_{\theta,M}^{(t)} \circ \mathcal{F}_{\theta,M-1}^{(t)} \circ \cdots \circ \mathcal{F}_{\theta,1}^{(t)} \circ \bm{x}_t.
\end{equation}
Unlike global rewards used in text-to-image models~\cite{fan2023dpok,black2024training} (e.g., CLIP score), physics-based rewards—such as PDE residuals—are inherently local, typically involving only neighboring points via finite difference schemes (see Algorithm~\ref{alg:residual_comp} for an example).

Motivated by this locality, we restrict parameter updates to only the final $m$ layers corresponding to the highest-resolution stage, i.e., $\{\mathcal{F}_{\theta,M}, \mathcal{F}_{\theta,M-1}, \ldots, \mathcal{F}_{\theta,M-m+1}\}$. In U-Net architectures, lower-resolution layers capture global, low-frequency semantics, while higher-resolution layers encode local, high-frequency details. By freezing the low-resolution layers, we constrain the model’s global semantics while allowing sufficient flexibility to satisfy localized physical constraints. Empirically, this strategy stabilizes training, mitigates reward hacking, and improves sample efficiency.

\noindent \textbf{Weight regularization (WR).} 
Conventional text-to-image reward fine-tuning methods~\cite{fan2023dpok,black2024training,uehara2024understanding} use distillation-based regularization to prevent reward hacking, typically via a KL divergence between the current and base model policies:
\begin{equation}
\label{equ:distill}
\max_{\theta} \mathbb{E} \left[ r(\bm{x}_0) - \lambda_{\text{distill}} \sum_{t=0}^{N-1} \| D_\theta(\bm{x}_t, t) - D_{\theta}^{\text{base}}(\bm{x}_t, t) \|_2^2 \right],
\end{equation}
However, this regularization doubles the number of forward passes, significantly reducing sample efficiency. Therefore, we re-examine the need for distillation under physics-based rewards. Noting that distillation objectives can be approximated under mild assumptions, we propose a weight-space alternative. If $D_\theta$ is Lipschitz-continuous with respect to $\theta$ (this is reasonable due to the normarlized inputs in practice), the KL term can be relaxed to a simple weight penalty:
\begin{equation}
\label{equ:weight}
\max_{\theta} \mathbb{E} \left[ r(\bm{x}_0) - \lambda_{\text{weight}} \| \theta - \theta_{\text{base}} \|_2^2 \right].
\end{equation}
We call this \textit{online weight regularization} (ON-WR). To further improve efficiency, we explore two lightweight, \textit{offline} strategies (OFF-WR). First, we observe that \textit{early stopping} can effectively limit weight drift, thereby mitigating reward hacking. Second, we apply \textit{linear interpolation} between the base and fine-tuned model, implemented as an exponential moving average (EMA) during fine-tuning. This implicit regularization approximates Equation~\eqref{equ:weight} post-hoc and offers additional sample efficiency.

\section{Experiments}
\subsection{Experimental setup}


\noindent \textbf{Datasets.} We validate our approach on five PDE datasets following~\cite{huang2024diffusionpde,shu2023physics}, including Burgers’ equation, Darcy flow, the inhomogeneous Helmholtz equation, the Poisson equation, and Kolmogorov flow. To ensure a fair comparison, we use the publicly available datasets provided in~\cite{huang2024diffusionpde,shu2023physics}. We focus on the unconditional setting as~\cite{bastek2025physicsinformed}, where the goal is to generate the entire physical fields, including both the PDE solution field and coefficient field. For example, in Darcy flow, we generate the concatenation of permeability (coefficient) and pressure (solution) fields. This setting can be naturally extended to conditional generation by training a conditional model. Detailed descriptions of each PDE and the data preparation process are provided in Appendix~\autoref{sec:benchmark+}, and a summary of the datasets is presented in~\autoref{tab:dataset}.


    

\noindent \textbf{Implementation details.}
We implement PIRF on top of the EDM~\cite{karras2022elucidating} framework. For consistency, we use base models from DiffusionPDE for most datasets. For Kolmogorov flow, which differs in its PDE setup, we train a base model from scratch using the official EDM configurations. We use a learning rate of 0.001 for pretraining and 0.0001 for fine-tuning. Fine-tuning is initially performed using 80 sampling steps, and we observe that the model generalizes well to 40-step inference. However, performance drops at 20 steps, so we fine-tune a separate model under the 20-step setting. We follow DRaFT~\cite{clark2024directly} to truncate the last step for 80-step sampling and the last two steps for 20-step sampling to save memory. Fine-tuning on approximately 180k trajectories is sufficient for strong reward optimization, using around 15h for 80-step fine-tuning and 5h for 20-step fine-tuning with two NVIDIA 80GB A100 GPUs. The EMA half-life is set as 50000.


\noindent \textbf{Baselines.}
We compare PIRF with several state-of-the-art physics-informed diffusion models, including DiffusionPDE~\cite{huang2024diffusionpde}, CoCoGen~\cite{jacobsen2025cocogen}, PG-Diffusion~\cite{shu2023physics}, and PIDM~\cite{bastek2025physicsinformed}. All models are implemented using the EDM~\cite{karras2022elucidating} framework for consistency. DiffusionPDE applies DPS~\cite{chung2022diffusion} to guide the reverse process, while CoCoGen further adds post-sampling correction steps. PG-Diffusion adopts classifier-free guidance (CFG)~\cite{ho2021classifier}, where the gradient of the PDE loss serves as the conditioning signal. Closer to our approach, PIDM incorporates a physics-informed supervision loss on the denoised output during training. For DiffusionPDE and CoCoGen, we follow the standard setup from DiffusionPDE by applying physics guidance during only the last 20\% of sampling steps. We perform grid searches over guidance scale and the number of correction steps and report the best results. For PG-Diffusion and PIDM, since their original papers do not cover all benchmarks used here, we extend their implementation to the new datasets using their official configurations. For PIDM, we use the mean estimation mode and initialize from the base model when training from scratch fails to converge due to instability.

\noindent \textbf{Evaluation metrics.}
All methods are evaluated using the Heun sampler with the default schedule from EDM using 20, 40, and 80 steps. We generate 1600 samples with same random seeds and compute the average PDE residual MSE as the quantitative metric. Furthermore, we visually checked whether samples are physcially plausible and diversified, in case of reward hacking happens in PIRF. We also compare their inference efficiency defined by number of reward function querying (NRQ), and number of backpropagating the model for gradient computation (NBM).

\begin{table}[t]
  \caption{Comparison of PDE residual mean squared error (MSE $\downarrow$) across different sampling steps.}
  \label{tab:pde-benchmark-merged}
  \centering
  \small
  \begin{tabular}{llccccc}
    \toprule
    \# Steps & Method & Burgers & Darcy  & Helmholtz & Poisson & Kolmogorov \\
    \midrule
    & Training data & 1.06E-02 & 16.44  & 1.10 & 1.20 & 14.00 \\
    \midrule
    \multirow{6}{*}{80} 
    & EDM~\cite{karras2022elucidating} & 2.01E-01 & 18.34  & 2.88 & 4.84 & 206.16 \\
    & DiffusionPDE~\cite{huang2024diffusionpde} & 8.38E-03 & 7.99  & 2.14 & 1.21 & \underline{51.04} \\
    & CoCoGen~\cite{jacobsen2025cocogen} & \underline{4.62E-03} & \underline{4.56}  & \underline{0.78} & \underline{0.95} & 76.66 \\
    & PG-Diffusion~\cite{shu2023physics} & 1.45E-01 & 7.97  & 7.81 & 8.65 & 408.11 \\
    & PIDM~\cite{bastek2025physicsinformed} & 1.67E-01 & 14.23  & 1.33 & 2.87 & 135.20 \\
    & \textbf{Ours} & \textbf{1.68E-03} & \textbf{1.29}  &\textbf{ 0.17 }& \textbf{0.19} & \textbf{19.28} \\
    \midrule
    \multirow{6}{*}{40}
    & EDM~\cite{karras2022elucidating} & 2.08E-01 & 25.43 & 9.71  & 5.72 & 205.26 \\
    & DiffusionPDE~\cite{huang2024diffusionpde} & 2.26E-02 & 22.73  & 8.71 & 3.02 & 84.53 \\
    & CoCoGen~\cite{jacobsen2025cocogen} & \underline{8.69E-03} & 8.25 & 1.63  & \underline{1.58} & \underline{80.32} \\
    & PG-Diffusion~\cite{shu2023physics} & 1.45E-01 & \underline{8.00} & 7.87  &  8.48& 368.39 \\
    & PIDM~\cite{bastek2025physicsinformed} & 1.73E-01 & 13.13  & \underline{1.31} & 2.85 & 136.79 \\
    & \textbf{Ours} & \textbf{2.31E-03} & \textbf{1.70} & \textbf{0.18}  & \textbf{0.20} & \textbf{19.37} \\
    \midrule
    \multirow{6}{*}{20}
    & EDM~\cite{karras2022elucidating} & 2.18E-01 & 28.94  & 5.20 & 5.07 & 219.40 \\
    & DiffusionPDE~\cite{huang2024diffusionpde} &5.60E-02&15.65&2.16&2.11&125.52 \\
    & CoCoGen~\cite{jacobsen2025cocogen} &1\textbf{.51E-02}&\underline{7.16}&\underline{0.75}&\underline{1.04}& \underline{101.27}\\
    & PG-Diffusion~\cite{shu2023physics} &1.32E-01&10.42&7.51&7.90&299.34\\
    & PIDM~\cite{bastek2025physicsinformed} &1.83E-01&9.45&1.19&2.56&143.84 \\
    & \textbf{Ours }&\underline{3.75E-02}&\textbf{1.99}&\textbf{0.28}&\textbf{0.28}&\textbf{20.62} \\
    \bottomrule
  \end{tabular}
\end{table}


\begin{table}[t]
    \centering
    \small
    \begin{minipage}[t]{0.48\linewidth}
        \centering
        \caption{Inference efficiency of different methods using Euler solver with $N$ steps and optional $M$ post-corrections. NRQ: number of reward queries. NBM: number of backpropagating model to compute reward gradient.}
        \begin{tabular}{llll}
        \toprule
        Method  & NRQ $\downarrow$ & NBM $\downarrow$\\
        \midrule
        PG-Diffusion~\cite{shu2023physics}           & $N$       &  $N$   \\
        DiffusionPDE~\cite{huang2024diffusionpde}     & $N$       &   $N$   \\
        CoCoGen~\cite{jacobsen2025cocogen}           & $N+M$     &   $N$   \\
        PIDM~\cite{bastek2025physicsinformed}       & $0$         &   $0$   \\
        Ours                                          & $0$         &     $0$ \\
        \bottomrule
        \end{tabular}
        \label{tab:time}
    \end{minipage}
    \hfill
    \begin{minipage}[t]{0.48\linewidth}
        \centering
        \caption{Sample efficiency analysis. $T$: total sample steps, $K$: steps truncated. Time: used seconds for fine-tuning 1k trajectories. LT: layer-wise truncation; WR: weight regularization. ON: online, OFF: offline.}
        \begin{tabular}{lcccc}
        \toprule
        $T$ & $K$ & LT & WR & Time (s) \\
        \midrule
        80 & 2 &         &  OFF       & 331 \\
        80 & 1 &         &   OFF      & 180 \\
        \midrule
        20 & 2 &         & ON & 163 \\
        20 & 2 &         &  OFF      & 87 \\
        20 & 2 & \checkmark &  OFF     & 84 \\
        \bottomrule
        \end{tabular}
        \label{tab:sample-efficiency}
    \end{minipage}    
\end{table}



\subsection{Results}
\noindent \textbf{Main results. } 
\autoref{tab:pde-benchmark-merged} compares PIRF with baseline methods across five PDE benchmarks using 80, 40, and 20 sampling steps. PIRF consistently outperforms all baselines in terms of physical precision, except on the Burgers' equation with 20 steps, where it slightly trails behind CoCoGen~\cite{jacobsen2025cocogen}. This may be attributed to the simplicity of the 1D Burgers' equation, where CoCoGen's post-correction procedure is particularly effective. However, on more complex PDEs, PIRF achieves significantly better performance than all baselines. Furthermore, PIRF not only surpasses baseline models but also matches or exceeds the physical precision of the training data itself, which serves as a “gold standard” derived from conventional simulators. In terms of inference efficiency, we evaluate methods based on the number of reward function queries (NRQ) and the number of model backward passes (NBM), as shown in~\autoref{tab:time}. PIRF and PIDM~\cite{bastek2025physicsinformed} are the most efficient at inference time, but PIRF achieves superior physical precision. Although CoCoGen is the second-best method in terms of physical enforcement across most benchmarks, it incurs the highest inference cost among all evaluated methods.
 

\begin{figure}[t]
    \centering
    \includegraphics[width=0.8\linewidth]{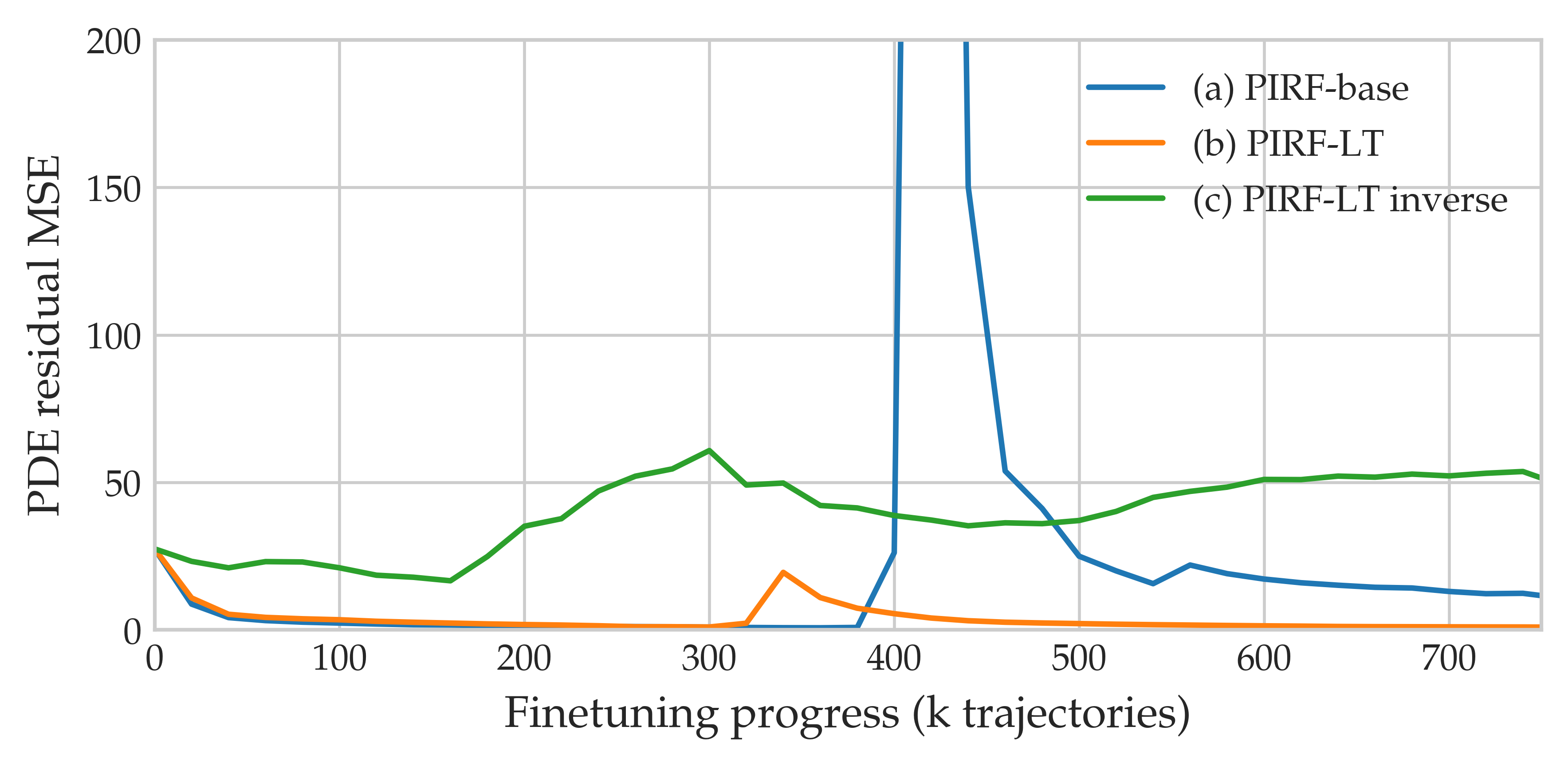}
    \caption{Effect of layer-wise truncation (LT). PIRF-LT (only updating high-resolution layer) achieves more stable long-term training compared to PIRF-base. As a inverse design, variant (c) fails to converge by only updating low-resolution layers, validating our assertion on the localized property of physics rewards. Additionally, PIRF-base exhibits signs of reward hacking while PIRF stays stable, as further illustrated in~\autoref{fig:layer-example}}.
    \label{fig:layerwise-plot}
\end{figure}



\noindent \textbf{Effect of layer-wise truncation. } 
We investigate the impact of layer-wise truncation (LT) in PIRF on the Darcy flow dataset. In our method, only the high-resolution decoder layers are updated during fine-tuning (PIRF-LT). We compare this against two baselines: (i) updating all layers (PIRF-base) and (ii) updating only the low-resolution decoder layers (PIRF-LT inverse). As shown in~\autoref{fig:layerwise-plot}, PIRF-LT maintains stable performance over long-term training and is robust to minor oscillations. In contrast, PIRF-base exhibits significant fluctuations and, once degraded, fails to recover. To better understand this behavior, we visualize the output fields at the 600k training step in~\autoref{fig:layer-example}. We observe that variant (i), which updates all layers, produces unnatural void regions—an indication of reward hacking and deviation from the true data distribution. In contrast, PIRF with LT preserves both training stability and distributional fidelity. For further comparison, we evaluate variant (ii), where only low-resolution layers are updated. This variant fails to converge, supporting our hypothesis that physics-based rewards are predominantly local and are most effectively optimized through high-resolution features. These results suggest that updating only high-resolution decoder layers is not only sufficient but also preferable for achieving high physical fidelity and stable training.

\begin{figure}
    \centering
    \includegraphics[width=.9\linewidth]{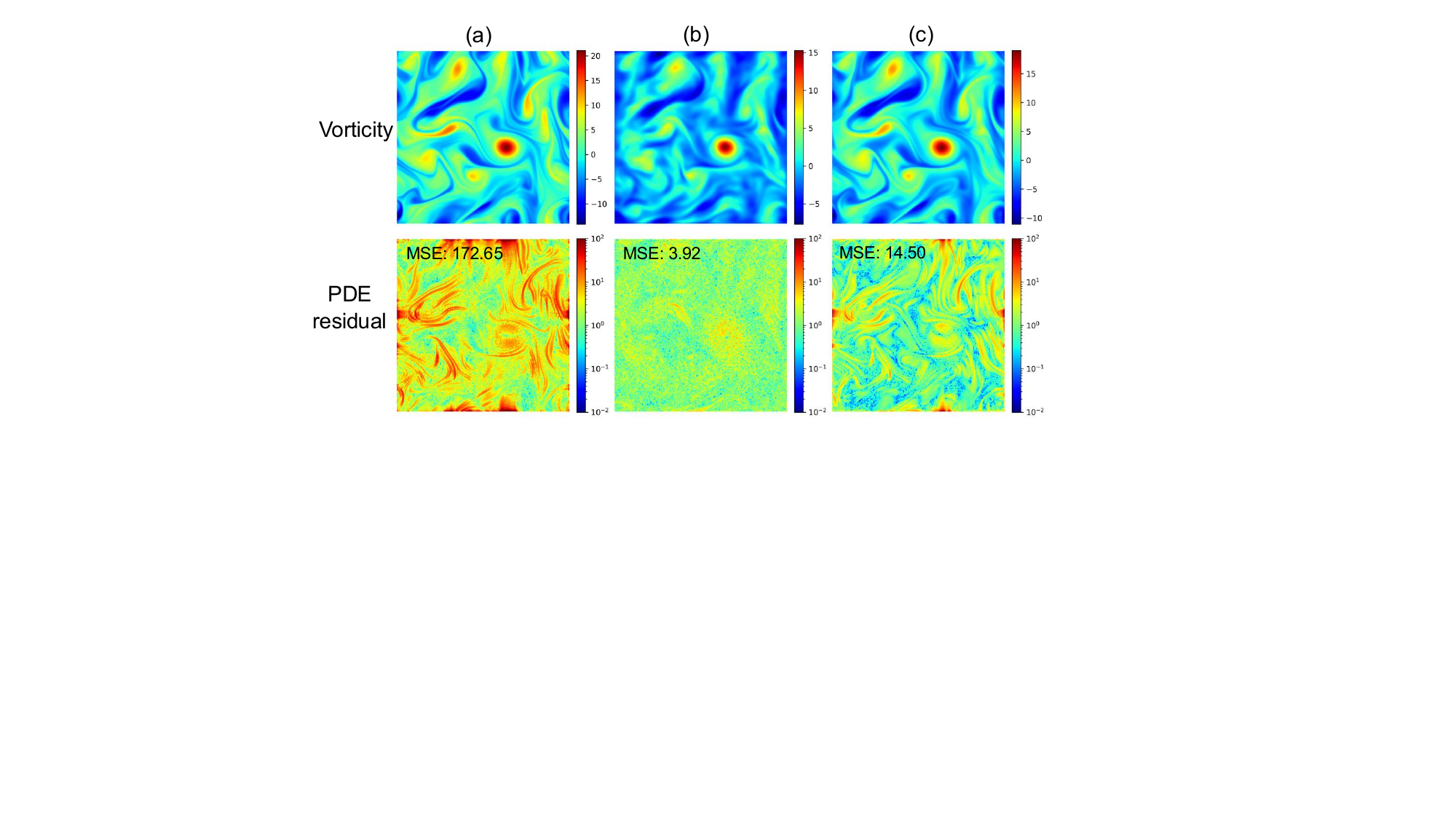}
    \caption{Effectiveness of weight regularization in mitigating reward hacking. From left to right: (a) sample from the base model on Kolmogorov flow, (b) PIRF without weight regularization, and (c) PIRF with offline weight regularization. While (b) achieves lowest PDE residual MSE, it introduces distortions in the vorticity field, indicating reward hacking. In contrast, (c) preserves structural consistency with the base model while maintaining low PDE residual.}
    \label{fig:ema}
\end{figure}

\noindent \textbf{Effect of weight regularization.}
The impact of weight regularization (WR) in PIRF is illustrated in~\autoref{fig:ema}. PIRF without WR (column b) achieves higher physical reward but significantly deviates from the base model's distribution (column a), exhibiting clear signs of reward hacking. In contrast, PIRF with WR (column c) achieves a more favorable balance between physical enforcement and data fidelity. Moreover, offline WR enhances sample efficiency compared to online WR. As shown in~\autoref{tab:sample-efficiency}, compared to online WR, offline WR reduces the fine-tuning time for 1,000 trajectories from approximately 163 seconds to 87 seconds—a nearly 2× improvement.

\section{Conclusion}

In this paper, we proposed PIRF, a physics-informed reward fine-tuning framework for diffusion models. Validated on five PDE benchmarks, PIRF achieves high physical precision under efficient sampling regimes. The introduction of layer-wise truncation and weight regularization further enhances both performance and training stability. We hope this work highlights the potential of reward fine-tuning as a principled approach to incorporating physical priors into generative models, bridging the gap between data-driven and model-based approaches in scientific domains.

\section*{Acknowledgements}
This work was supported by the National Science Foundation AI Institute (NSF Award No. 2112085) and by the National Institutes of Health (NIH Award No. R01HL159183).
\newpage
\bibliographystyle{unsrt}
\bibliography{refs}


\newpage
\appendix

\section{Technical Appendices and Supplementary Material}
\subsection{More on methodology}
\label{sec:method+}

\begin{figure}[t]
    \centering
    \includegraphics[width=0.7\linewidth]{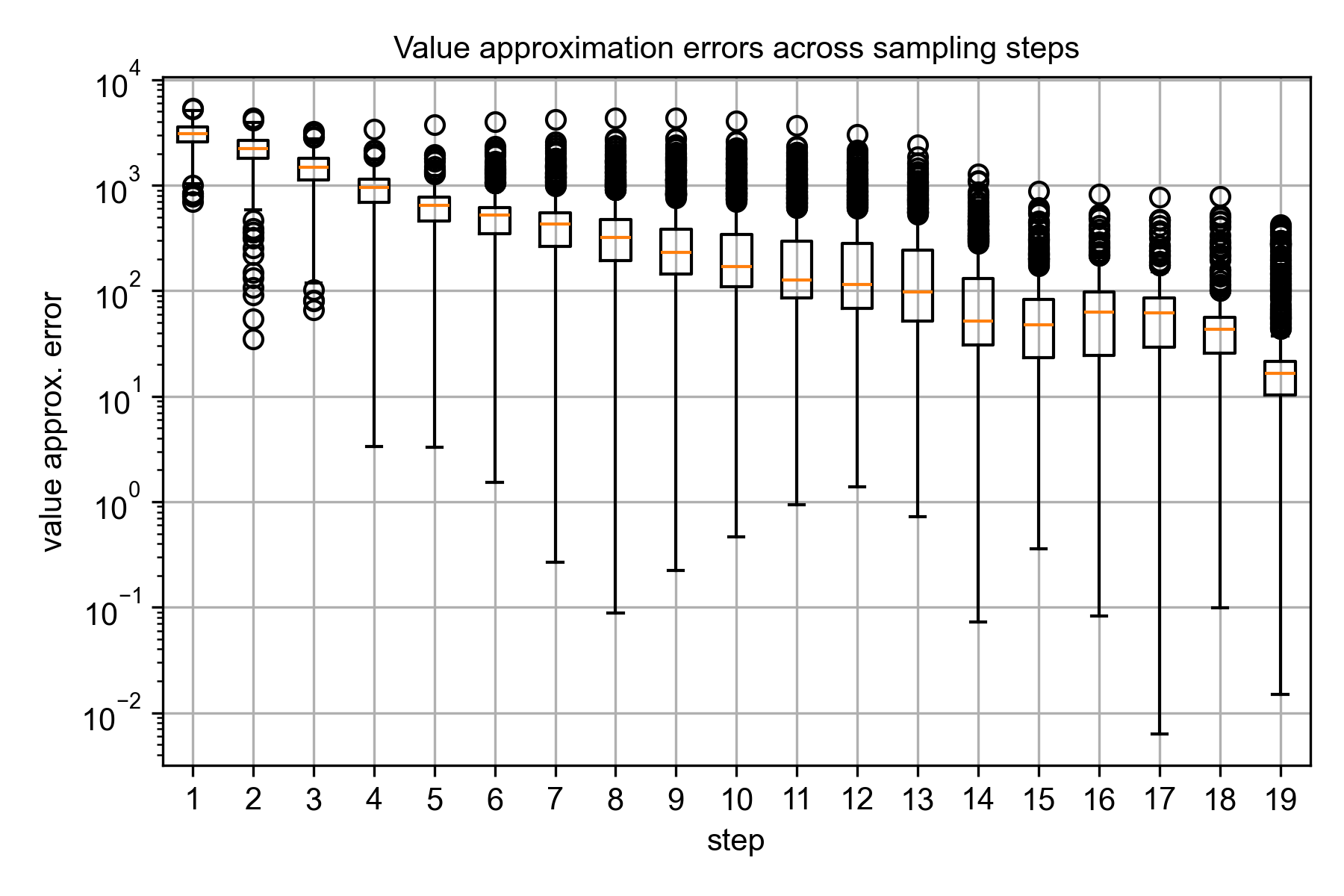}
    \caption{Box plot showing the distribution of value approximation errors across sampling steps in existing approaches. Notably, the approximation error remains high even in late sampling steps, highlighting a key bottleneck in current physics-informed diffusion models.}
    \label{fig:value}
\end{figure}

\noindent \textbf{Analysis on value function approximation errors} \autoref{fig:value} shows the distribution of value approximation errors across sampling steps. Given a trajectory $\{\bm{x}_t\}_{t=T}^{0}$, the $y$-axis (log scale) represents the absolute error of value approximation, defined as $|r(\hat{\bm{x}}_{0}^{t}) - r(\bm{x}_0)|$, where $\hat{\bm{x}}_{0}^{t} = D_\theta(\bm{x}_0,t)$ is the point estimate of final state given $\bm{x}_t$. The $x$-axis corresponds to the step index $t+1$, with $T=20$ chosen to reflect efficient sampling settings. We use the base model from DiffusionPDE~\cite{huang2024diffusionpde} for Darcy flow to plot this graph with 800 random samples. We can find that value function approxmation errors remains high even in late steps, leading to a key bottleneck for existing approaches. Motivated by this, we use physics-informed reward backpropagation to bypass this bottleneck.

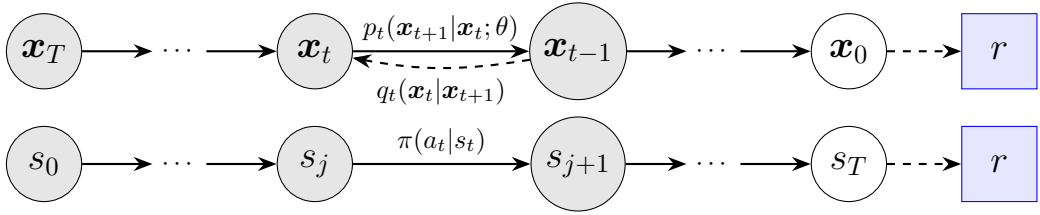
\begin{figure}[ht]
\centering
\begin{tikzpicture}[node distance=2cm, on grid, auto]
    \tikzset{
        state/.style={circle, draw=black, fill=gray!20, minimum size=1cm, font=\Large\bfseries},
        visible/.style={circle, draw=black, minimum size=1cm, font=\Large\bfseries},
        reward/.style={rectangle, draw=blue, fill=blue!10, minimum height=1cm, minimum width=1cm, font=\Large\bfseries},
        arrow/.style={-{Stealth[length=2.5mm]}, thick},
        dashed_arrow/.style={-{Stealth[length=2.5mm]}, thick, dashed},
        dotted_box/.style={draw=black, dotted, very thick, inner sep=10pt, rounded corners}
    }

    \node[state] (xT) {$\bm{x}_T$};
    \node[right=1.8cm of xT] (dots1) {$\cdots$};
    \node[state, right=1.8cm of dots1] (xt) {$\bm{x}_t$};
    \node[state, right=3.5cm of xt] (xt1) {$\bm{x}_{t-1}$};
    \node[right=1.8cm of xt1] (dots2) {$\cdots$};
    \node[visible, right=1.8cm of dots2] (x0) {$\bm{x}_0$};
    \node[reward, right=2cm of x0] (r) {$r$};

    \draw[arrow] (xT) -- (dots1);
    \draw[arrow] (dots1) -- (xt);
    \draw[arrow] (xt) -- node[above] {$p_{t}(\bm{x}_{t+1} | \bm{x}_t;\theta)$} (xt1);
    \draw[arrow] (xt1) -- (dots2);
    \draw[arrow] (dots2) -- (x0);
    \draw[dashed_arrow] (x0) -- (r);

    \draw[dashed_arrow, bend left=10] (xt1) to node[below] {$q_t(\bm{x}_t |\bm{x}_{t+1})$} (xt);

    \node[state, below=1.5cm of xT] (s1) {$s_0$};
    \node[below=1.5cm of dots1] (dots3) {$\cdots$};
    \node[state, below=1.5cm of xt] (st) {$s_{j}$};
    \node[state, below=1.5cm of xt1] (st1) {$s_{j+1}$};
    \node[below=1.5cm of dots2] (dots4) {$\cdots$};
    \node[visible, below=1.5cm of x0] (sT) {$s_{T}$};
    \node[reward, below=1.5cm of r] (r2) {$r$};

    \draw[arrow] (s1) -- (dots3);
    \draw[arrow] (dots3) -- (st);
    \draw[arrow] (st) -- node[above] {$\pi(a_t | s_{t})$} (st1);
    \draw[arrow] (st1) -- (dots4);
    \draw[arrow] (dots4) -- (sT);
    \draw[dashed_arrow] (sT) -- (r2);
\end{tikzpicture}
\caption{Formulating diffusion model fine-tuning as a sequential decision-making problem with sparse reward. Here $j+t=T$, The top row illustrates the reverse generative process; the bottom row shows its equivalent MDP.}
\label{fig:graphic}
\end{figure}


\begin{figure}[t]
    \centering
    \includegraphics[width=\linewidth]{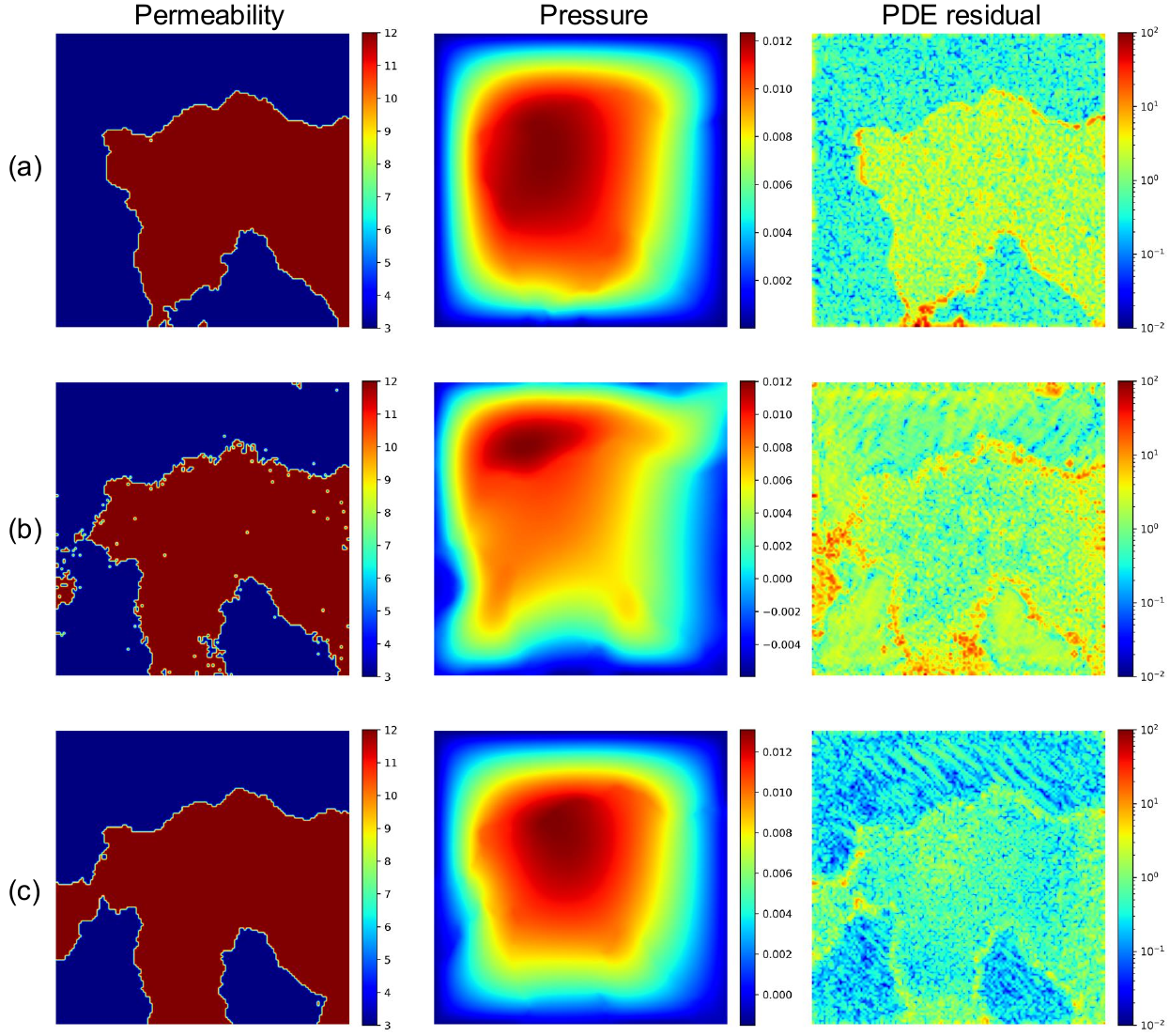}
    \caption{Illustrative example at the 600k fine-tuning checkpoint on Darcy flow. From top to bottom: (a) base model, (b) PIRF without layer-wise truncation, (c) PIRF with layer-wise truncation. In (b), the permeability field exhibits artificial voids within originally intact regions, reflecting a divergence from the data distribution driven by over-optimization of the physics reward—an instance of reward hacking. In contrast, (c) shows that layer-wise truncation preserves structural integrity, yielding more stable and physically consistent outputs.}
    \label{fig:layer-example}
\end{figure}

\begin{algorithm}[ht]
\label{alg:residual_comp}
\caption{Pseudocode of computing Burgers' equation residual in a PyTorch-like style.}
\begin{lstlisting}[language=Python]
class BurgersResidual(nn.Module):
    def __init__(self, domain_length=1.0, pixels_per_dim=128):
        super().__init__()
        dx = domain_length / pixels_per_dim
        dt = domain_length / pixels_per_dim
        self.deriv_x = tensor([[-1, 0, 1]]).view(1,1,1,3) / (2 * dx)
        self.deriv_t = tensor([[-1], [0], [1]]).view(1,1,3,1) / (2 * dt)

    def forward(self, u):
        # u: input of shape (B, 1, T, X)
        # u partial on x
        u_x = pad(u, (1,1,0,0), mode='replicate')
        u_x = conv2d(u_x, self.deriv_x)

        # u partial on t
        u_t = pad(u, (0,0,1,1), mode='replicate')
        u_t = conv2d(u_t, self.deriv_t)

        # u_x partial on x
        u_xx = pad(u_x, (1,1,0,0), mode='replicate')
        u_xx = conv2d(u_xx, self.deriv_x)

        # compute residual
        return u_t + u * u_x - 0.01 * u_xx
\end{lstlisting}
\end{algorithm}

\subsection{Details on benchmarks}
\label{sec:benchmark+}
A summary of benchmarks is provided in~\autoref{tab:dataset}. Here we illustrate dataset simulation details for each PDE. For each PDE, if not specified, we generate $50,000$ samples.
\begin{table}[b]
    \centering

    \small
    \caption{Summary of PDE datasets.}
    
    \label{tab:dataset}
    \begin{tabular}{lcccc}
    \toprule
        Dataset & Spatial resolution & Temporal resolution & \# Samples & Source \\
    \midrule
        Burgers' equation & $128$ & $128$ & 50{,}000 & \cite{huang2024diffusionpde} \\
        Darcy flow         & $128 \times 128$ & N/A & 50{,}000 & \cite{huang2024diffusionpde} \\
        Helmholtz equation & $128 \times 128$ & N/A & 50{,}000 & \cite{huang2024diffusionpde} \\
        Poisson equation   & $128 \times 128$ & N/A & 50{,}000 & \cite{huang2024diffusionpde} \\
        Kolmogorov flow    & $256 \times 256$ & $320$ & 40 & \cite{shu2023physics} \\
    \bottomrule
    \end{tabular}
\end{table}

\subsubsection{Darcy flow}

Darcy flow equations~\cite{jacobsen2025cocogen} describe the relationship between fluid pressure and the permeability of a porous medium. A coefficient field \( \bm{a}(\boldsymbol{\xi}) \) characterizes the ease of fluid flow at each spatial location \( \boldsymbol{\xi} \in \Omega \), while a source function \( f(\boldsymbol{\xi}) \) represents fluid injection or extraction. The pressure field \( \bm{u}(\boldsymbol{\xi}) \) and velocity field \( \boldsymbol{v}(\boldsymbol{\xi}) \) satisfy:
\begin{equation}
\label{equ:darcy}
    \begin{aligned}
        \boldsymbol{v}(\boldsymbol{\xi}) &= -\bm{a}(\boldsymbol{\xi}) \nabla \bm{u}(\boldsymbol{\xi}), \quad \boldsymbol{\xi} \in \Omega, \\
        \nabla \cdot \boldsymbol{v}(\boldsymbol{\xi}) &= f(\boldsymbol{\xi}), \quad \boldsymbol{\xi} \in \Omega, \\
        \boldsymbol{v}(\boldsymbol{\xi}) \cdot \hat{\boldsymbol{n}}(\boldsymbol{\xi}) &= 0, \quad \boldsymbol{\xi} \in \partial \Omega, \\
        \int_{\mathcal{X}} \bm{u}(\boldsymbol{\xi}) \, \mathrm{d} \boldsymbol{\xi} &= 0.
    \end{aligned}
\end{equation}

\begin{equation}
    f(\boldsymbol{\xi}) =
\begin{cases} 
r, & \left| \xi_i - \frac{1}{2} w \right| \leq \frac{1}{2} w, \quad i = 1, 2, \\ 
-r, & \left| \xi_i - 1 + \frac{1}{2} w \right| \leq \frac{1}{2} w, \quad i = 1, 2, \\ 
0, & \text{otherwise}.
\end{cases}
\end{equation}


We follow the settings from~\cite{huang2024diffusionpde}, a constant source is used: \( f(\boldsymbol{\xi}) = 1 \). We first generate Gaussian random fields on \( (0, 1)^2 \) with \( \mu \sim \mathcal{N}(0, (\Delta + 9\bm{I})^{-2}) \), and then define:
\begin{equation}
\bm{a}(\xi) =
    \begin{cases}
        12, & \text{if } \mu(\xi) \geq 0,\\
        3, & \text{if } \mu(\xi) < 0.
    \end{cases}
\end{equation}

The equations are solved using finite difference approximations on a spatial domain \( \mathcal{X} = [0, 1]^2 \) with an \( n \times n \) grid~\cite{jacobsen2025cocogen} (where \( n = 128 \)). Each grid point indexed by \( i, j \) corresponds to:
\begin{equation}
    \boldsymbol{\xi}_{i,j} = \left[\frac{i-1}{n-1}, \frac{j-1}{n-1}\right]^\top, \quad i, j = 1, \dots, n.
\end{equation}
The solution \( \bm{u}(\boldsymbol{\xi}) \) is vectorized as \( \boldsymbol{u} \in \mathbb{R}^{n^2} \), yielding a linear system \( \boldsymbol{A} \boldsymbol{u} = \boldsymbol{f} \), where \( \boldsymbol{A} \in \mathbb{R}^{(n^2 + 1) \times n^2} \). The final row enforces the integral constraint \( \int_{\mathcal{X}} \bm{u}(\boldsymbol{\xi}) \, \mathrm{d} \boldsymbol{\xi} = 0 \). Derivatives are approximated using finite differences, and boundary conditions are applied by adjusting stencils. The velocity is recovered as \( \boldsymbol{v}(\boldsymbol{\xi}) = -\bm{a}(\boldsymbol{\xi}) \nabla \bm{u}(\boldsymbol{\xi}) \), with gradients estimated using second-order central differences.

To assess physical consistency, we compute the residual:
\begin{equation}
\label{equ:darcy-residual}
    \bm{\mathcal{R}}(\bm{u}, \bm{a}) = \bm{a}(\boldsymbol{\xi}) \frac{\partial^2 \bm{u}(\boldsymbol{\xi})}{\partial \xi_1^2} 
    + \frac{\partial \bm{a}(\boldsymbol{\xi})}{\partial \xi_1} \frac{\partial \bm{u}(\boldsymbol{\xi})}{\partial \xi_1} 
    + \bm{a}(\boldsymbol{\xi}) \frac{\partial^2 \bm{u}(\boldsymbol{\xi})}{\partial \xi_2^2} 
    + \frac{\partial \bm{a}(\boldsymbol{\xi})}{\partial \xi_2} \frac{\partial \bm{u}(\boldsymbol{\xi})}{\partial \xi_2} 
    + f(\boldsymbol{\xi}).
\end{equation}
The residual is evaluated at each grid point. The integral constraint is enforced by normalizing:
\(\boldsymbol{u} = \tilde{\boldsymbol{u}} - \int_{\mathcal{X}} \tilde{\boldsymbol{u}} \, \mathrm{d} \boldsymbol{\xi},\)
ensuring the residual reflects only PDE compliance.

\subsubsection{Inhomogeneous Helmholtz equation}

We consider the static inhomogeneous Helmholtz equation with no-slip boundary conditions on \( \partial \Omega \), describing wave propagation:
\begin{equation}
\label{equ:helm}
\begin{aligned}
    \nabla^2 \bm{u}(\bm{\xi}) + k^2 \bm{u}(\bm{\xi}) &= \bm{a}(\bm{\xi}), \quad \bm{\xi} \in \Omega, \\
    \bm{u}(\bm{\xi}) &= 0, \quad \bm{\xi} \in \partial \Omega.
\end{aligned}
\end{equation}
Here, \( k \) is a constant. When \( k = 0 \), \autoref{equ:helm} reduces to the Poisson equation. We set \( k = 1 \) for the Helmholtz case. The residual is computed as:
\begin{equation}
    \bm{\mathcal{R}}(\bm{u}, \bm{a}) = \nabla^2 \bm{u}(\bm{\xi}) + k^2 \bm{u}(\bm{\xi}) - \bm{a}(\bm{\xi}).
\end{equation}

Following~\cite{huang2024diffusionpde}, we first generate Gaussian fields on \( (0, 1)^2 \) with \( \bm{a} \sim \mathcal{N}(0, (\Delta + 9\bm{I})^{-2}) \), then solve for \( \bm{u} \) using second-order finite differences. To enforce the no-slip boundary condition, we multiply the solution by a mollifier \( \sin(\pi\xi_1)\sin(\pi\xi_2) \), where \( \bm{\xi} = (\xi_1, \xi_2) \in (0,1)^2 \). Both \( \bm{a} \) and \( \bm{u} \) are defined on \( 128 \times 128 \) grids.
\subsubsection{Kolmogorov flow}
\label{subsec:kol}

The dataset used in this study is based on the two-dimensional Kolmogorov flow~\cite{chandler2013invariant}, governed by the incompressible Navier–Stokes equations in vorticity form:
\begin{equation}
\label{equ:kol}
    \begin{aligned}
        \frac{\partial \bm{u}(\bm{\xi}, \tau)}{\partial \tau} + \bm{\nu}(\bm{\xi}, \tau) \cdot \nabla \bm{u}(\bm{\xi}, \tau) &= \frac{1}{Re} \nabla^2 \bm{u}(\bm{\xi}, \tau) + f(\bm{\xi}), \quad \bm{\xi} \in (0,2\pi)^2, \, \tau \in (0,T], \\
        \nabla \cdot \bm{\nu}(\bm{\xi}, \tau) &= 0, \quad \bm{\xi} \in (0,2\pi)^2, \, \tau \in (0,T], \\
        \bm{u}(\bm{\xi}, 0) &= \bm{u}_0(\bm{\xi}), \quad \bm{\xi} \in (0,2\pi)^2.
    \end{aligned}
\end{equation}

Here, \(\bm{u}\) denotes the vorticity field, \(\bm{\nu}\) is the velocity field, and \(Re\) is the Reynolds number, set to 1000. The function \(f(\bm{\xi})\) represents the external forcing term, and \(\bm{\xi} = [\xi_1, \xi_2]\) is the spatial coordinate. We follow the settings in~\cite{shu2023physics}, where periodic boundary conditions are applied. The forcing used for the 2D Kolmogorov flow is defined as:
\(
f(\bm{\xi}) = -4\cos(4\xi_2) - 0.1\, \bm{u}(\bm{\xi}, \tau).
\)

To numerically solve~\autoref{equ:kol}, we use the pseudo-spectral solver from~\cite{li2024physics}. The initial condition \(\bm{u}_0(\bm{\xi})\) is sampled from a Gaussian random field:
\(
\bm{u}_0(\bm{\xi}) \sim \mathcal{N}\left(0, 7^{3/2}(-\Delta + 49I)^{-5/2}\right).
\)
Simulations are performed on a \(2048 \times 2048\) uniform grid. 
We generate 40 sequences, each simulating 10 seconds of dynamics (\(T=10\)). These are downsampled spatially to a \(256 \times 256\) grid and temporally using a fixed step size \(\Delta \tau = 1/32\) s, yielding 320 frames per sequence. Of these, 36 sequences are used for training and the remaining 4 for testing. 

For computing the PDE residual \(\bm{\mathcal{R}}(\bm{u})\), we follow~\cite{shu2023physics,li2025physicsaligned}, using the discrete Fourier transform to compute spatial derivatives and finite differences for time derivatives. As the proposed diffusion model operates on vorticity over three consecutive frames \([\bm{u}_{\tau-1}(\bm{\xi}), \bm{u}_\tau(\bm{\xi}), \bm{u}_{\tau+1}(\bm{\xi})]\), we approximate the time derivative as:
\(\partial_\tau \bm{u}(\bm{\xi}, \tau) \approx (\bm{u}_{\tau+1}(\bm{\xi}) - \bm{u}_{\tau-1}(\bm{\xi}))/{2\Delta \tau}.\)
The convection and diffusion terms are computed in Fourier space by estimating gradients and Laplacians of the vorticity. The velocity field is derived via the stream function \(\psi\) using:
\begin{equation}
    \bm{\nu} = \nabla \times \psi, \quad -\nabla^2 \psi = \bm{u},
\end{equation}
and transformed back to physical space using the inverse Fourier transform.

\subsubsection{Burgers' Equation}

We study the one-dimensional viscous Burgers' equation with periodic boundary conditions on a unit-length spatial domain \( \Omega = (0,1) \). The governing equation is:
\begin{equation}
\label{equ:burgers}
\begin{aligned}
\partial_\tau \bm{u}(\xi, \tau) + \partial_{\xi} \left(\frac{1}{2} \bm{u}^2(\xi, \tau)\right) &= \nu \, \partial_{\xi\xi} \bm{u}(\xi, \tau), \quad \xi \in \Omega,\, \tau \in (0, T], \\
\bm{u}(\xi, 0) &= \bm{u}_0(\xi), \quad \xi \in \Omega.
\end{aligned}
\end{equation}
The viscosity is set to \( \nu = 0.01 \). Following the setup in~\cite{huang2024diffusionpde,li2024physics}, the initial condition \( \bm{u}_0 \) is sampled from a Gaussian random field:
\(\bm{u}_0 \sim \mathcal{N}(0,\, 625(\Delta + 25\bm{I})^{-2}).
\) We simulate the system for 1 second using a spectral solver, discretizing both space and time. The spatial grid consists of 128 points, and we take 127 time steps after the initial state, resulting in a solution tensor \( \bm{u}_{0:T} \in \mathbb{R}^{128 \times 128} \).

Because the solution is modeled densely over time, the PDE residual can be reliably approximated using finite difference schemes:
\begin{equation}
    \bm{\mathcal{R}}(\bm{u}) = \partial_\tau \bm{u}(\xi, \tau) + \partial_\xi \left( \frac{1}{2} \bm{u}^2(\xi, \tau) \right) - \nu \, \partial_{\xi\xi} \bm{u}(\xi, \tau).
\end{equation}
This residual is evaluated at each point in the spatiotemporal domain to assess physical consistency.

\subsection{Related work}
\label{sec:related}
\paragraph{Reward fine-tuning}~\cite{uehara2024understanding} has been widely applied to adapt pre-trained text-to-image diffusion models to downstream reward functions, such as human preference alignment. 
Four major classes of paradigms have emerged in this space:
(i) \textit{Reward-weighted regression}~\cite{lee2023aligning,dong2023raft} collects samples and their associated rewards, and then fine-tunes the diffusion model using a reward-weighted loss;
(ii) \textit{Policy gradient} methods~\cite{fan2023dpok,black2024training,Deng_2024_CVPR} optimize the expected reward directly using reinforcement learning techniques such as proximal policy optimization~\cite{schulman2017proximal};
(iii) \textit{Reward backpropagation}~\cite{clark2024directly,wu2024deep,uehara2024feedback} assumes a differentiable reward function and directly propagates the reward signal through the diffusion process to update model parameters. This approach is training-efficient but suffers from challenges such as the depth-memory dilemma~\cite{wu2024deep} and reward hacking;
(iv) \textit{Value-weighted sampling}~\cite{uehara2024understanding,uehara2024feedback} avoids parameter updates by using an estimated value function to guide sampling, limited by the quality of the value function approximation. 


In this work, we extend reward fine-tuning to physics-informed generation. We focus on the reward backpropagation approach and further investigates a layer-wise truncation schedule beyond conventional step-wise truncation. We also re-examine the necessity of distillation-based regularization in the new context, and replace it with a more efficient yet effective weight regularization.

\paragraph{Physics-informed diffusion models} aim to incorporate physical constraints PDEs into diffusion models~\cite{huang2024diffusionpde,shu2023physics,bastek2025physicsinformed,li2025physicsaligned,jacobsen2025cocogen}. \textit{Guidance-based} methods typically build on the DPS framework~\cite{chung2022diffusion}, where physics-based guidance is applied during sampling using point estimates of the data distribution via Tweedie’s formula~\cite{efron2011tweedie}. \textit{Training-based} approaches, on the other hand, integrate physics constraints directly into the learning objective. The specific approaches are discussed in~\autoref{sec:prior}, with further details provided in~\autoref{sec:method+}. PalSB~\cite{li2025physicsaligned}, most closely related to ours, applies DRaFT~\cite{clark2024directly} in a diffusion bridge to improve physical fidelity. However, PalSB is tailored to field reconstruction tasks and the broader potential of reward fine-tuning for physics alignment remains underexplored. 

Our method advances this line of work in three key ways. First, we reinterpret existing methods from the reward optmization perspective and identify the key bottleneck from value approximation. Second, compared to PalSB, we investigate a layer-wise truncation strategy. Third, we show that data- or distillation-based regularization in PalSB is unnecessary and propose a more efficient weight-based regularization scheme.

\paragraph{Diffusion Models for PDEs}
Beyond the core studies discussed earlier, a growing body of work explores the use of diffusion models in the context of PDEs, primarily targeting PDE-solving tasks, with a particular emphasis on fluid dynamics and other time-evolving systems. For example, Kohl et al.~\cite{kohl2023benchmarking} introduced an autoregressive formulation for PDE simulation, while Yang and Sommer~\cite{yang2023denoising} proposed directly predicting future states. Cachay et al.~\cite{ruhling2023dyffusion} recently unified denoising time and physical time to improve scalability. Lippe et al.~\cite{lippe2023pde} built upon neural operators with iterative denoising refinement to enhance the modeling of high-frequency components and enable long-term rollouts. Other works focus on downstream tasks such as data assimilation and super-resolution. Shu et al.~\cite{shu2023physics} and Jacobsen et al.~\cite{jacobsen2025cocogen} enforce PDE constraints during inference to improve physical consistency. Huang et al.~\cite{huang2024diffusionpde} used an amortized diffusion model combined with inpainting~\cite{chung2022diffusion} to perform data assimilation on weather datasets. ECI~\cite{cheng2025gradientfree} proposed a training- and gradient-free framework for adapting pre-trained functional flow matching (FFM) models~\cite{kerrigan2024functional} to exactly satisfy boundary or initial conditions, rather than full PDE constraints. Rozet and Louppe~\cite{rozet2023score} decomposed long trajectories into short segments with local score estimation to improve memory efficiency, and Qu et al.~\cite{qu2024deep} extended this approach using latent diffusion models for data assimilation on ERA5 weather datasets. Ruhe et al.~\cite{ruhe2024rolling} further introduced a frame-dependent noising process for video and fluid dynamics generation using local scores. DiffPhyCon~\cite{wei2024diffphycon} addresses PDE control problems by training diffusion models over both state trajectories and control sequences, guided by the control objective. Shysheya et al.~\cite{shysheya2024conditional} conducted a comprehensive comparison of conditional diffusion models for PDE simulation tasks.

While these works share some overlap with ours in enforcing physical constraints, their primary focus lies in PDE solving, where the objective is to achieve high reconstruction or solution accuracy. In contrast, our approach centers on enforcing physical consistency intrinsically within the generative model. That is, we focus on the model’s inherent ability to produce samples that satisfy physical laws. Furthermore, our method can be naturally extended to conditional settings by adapting the generative model to incorporate conditioning inputs.

\subsection{Limitations and Future Directions}
In this work, we focus on the unconditional generation setting. However, PIRF can be readily extended to task-specific conditional settings, such as field reconstruction~\cite{li2025physicsaligned}. A more challenging direction is adapting a fine-tuned unconditional model to support a variety of conditioning types under few-step sampling—for example, solving forward or inverse PDE problems with sparse observations~\cite{huang2024diffusionpde}. In such cases, combining reward fine-tuning with additional control mechanisms~\cite{zhao2025adding,zhang2023adding} may offer a promising solution. Moreover, our current implementation uses deterministic sampling for fine-tuning. Exploring stochastic samplers may further improve performance and generalization. Finally, like prior works, we assume the physics-based reward is fully known and differentiable. Extending PIRF to settings with partially known or non-differentiable physics remains an exciting future direction, where policy gradient-based approaches~\cite{fan2023dpok,black2024training,Deng_2024_CVPR} could offer a viable alternative.

\end{document}